\documentclass[conference]{IEEEtran}
\IEEEoverridecommandlockouts
\IEEEpubid{\makebox[\columnwidth]{ 978-1-6654-3540-6/22 © 2022  IEEE \hfill} \hspace{\columnsep}\makebox[\columnwidth]{ }}

\usepackage{amsmath,amssymb,amsfonts}
\usepackage{algorithmic}
\usepackage{graphicx}
\usepackage{subcaption}
\usepackage{bbm}
\usepackage{booktabs}
\usepackage{textcomp}
\usepackage{xcolor}

\def\BibTeX{{\rm B\kern-.05em{\sc i\kern-.025em b}\kern-.08em
    T\kern-.1667em\lower.7ex\hbox{E}\kern-.125emX}}
    
\begin{document}

\title{Decentralized Coordination in Partially Observable Queueing Networks\\
\thanks{This work has been funded by the German Research Foundation (DFG) via the Collaborative Research Center (CRC) 1053 – MAKI.}
}

\author{\IEEEauthorblockN{Jiekai~Jia, Anam~Tahir, Heinz~Koeppl}
\IEEEauthorblockA{ \textit{Department of Electrical Engineering and Information Technology} \\
\textit{Technische Universität Darmstadt}\\
Darmstadt, Germany \\
{\tt\small zjsxsyjjk@gmail.com,
\tt\small \{anam.tahir, heinz.koeppl\}@tu-darmstadt.de}
}
}

\maketitle

\begin{abstract}
We consider communication in a fully cooperative multi-agent system, where the agents have partial observation of the environment and must act jointly to maximize the overall reward. We have a discrete-time queueing network where agents route packets to queues based only on the partial information of the current queue lengths. The queues have limited buffer capacity, so packet drops happen when they are sent to a full queue. In this work, we implemented a communication channel for the agents to share their information in order to reduce the packet drop rate. For efficient information sharing we use an attention-based communication model, called ATVC, to select informative messages from other agents. The agents then infer
the state of queues using a combination of the variational auto-encoder, VAE, and product-of-experts, PoE, model. Ultimately, the agents learn what they need to communicate and with whom, instead of communicating all the time with everyone.
We also show empirically that ATVC is able to infer the true state of the queues 
and leads to a policy which outperforms existing baselines.
\end{abstract}


\section{INTRODUCTION}
Deep reinforcement learning has been remarkably successful in a variety of complex single-agent problems, such as playing games~\cite{2013_DQN},~\cite{2016_alphago}, robotics control~\cite{2016_RC},~\cite{2017_RC} and video streaming control~\cite{2017_video_stream}. This progress encouraged the research on its multi-agent extensions, such as value-based methods~\cite{2017_IDQN},~\cite{2018_Qmix},~\cite{2019_Qtran} and policy gradient methods~\cite{2017_MADDPG}, ~\cite{2020_IPPO}, ~\cite{2021_MAPPO}. In a multi-agent system, the coordination and communication between agents is essential to achieve global goals, especially in a decentralized manner with partial observations.

There are many solutions already available in order to achieve and improve coordination between agents in a partially observable environment. One main approach is to implement a communication channel between agents for the purpose of information sharing, especially in a decentralized setup. 
DIAL~\cite{2016_DIAL}, CommNet~\cite{2016_CommNet}, BicNet~\cite{2017_bicnet}, Master-Slave~\cite{2017_master_slaver}, ATOC~\cite{2018_ATOC}, SchedNet~\cite{2019_schedulecomm} and NDQ~\cite{2019_commminimum} are a few of the recent significant models in this direction. 
However, the communication models mentioned above all assume that the agents have correct or updated information of the environment at all times, which may not be realistic.

In this paper, we consider homogeneous, fully cooperative schedulers (agents) which are sensing, communicating and acting in a partially observable queueing network, in order to minimize the total packet drop rate. The partial observability comes from the fact that the agents may sense noisy observations of the queue states, which makes the communication crucial in helping to infer the true state of the queues. 
Such a scenario can occur in large data centers or in manufacturing industries, and the schedulers should be capable of communicating efficiently and organizing themselves to improve the overall performance of the decentralized system.
To this end, we propose a communication model, called attention-based variational communication model~(ATVC), which consists of a multimodal variational autoencoder,~(MVAE)~\cite{2018_MVAE} and an attention module,~\cite{2014_attention},~\cite{2015_memory_network}. MVAE combines the observations from different agents for the inference of underlying true observations, and the attention module is used to pick out the most valuable information to share for efficient communication. Furthermore, the model is implemented under the paradigm of centralized learning and decentralized execution~\cite{2008_CTDE},~\cite{2016_CTDE}, which has proven to be an effective method in multi-agent reinforcement learning.

\section{RELATED WORK}
Recently, a lot of work has explored the importance of communication in multi-agent reinforcement learning, particularly differentiable communication \cite{2016_DIAL}, which has shown to improve the learning performance. 
The authors in~\cite{2016_CommNet} propose a differentiable communication network, CommNet. Compared to traditional communication methods, CommNet does not set communication actions to agents, it instead incorporates the communication channel in networks, and therefore, the gradient can flow across agents through the communication channel and not just within an agent. The experimental results show that differentiable communication significantly improves the performance. However, CommNet is a broadcast communication. This can increase the computational complexity with the growth in the number of agents. Interestingly, the authors in~\cite{2018_abst_summarise} propose an abstract summarization model based on CommNet, which proves that CommNet is a reliable model with wide applications.

Nearly simultaneously, the authors in~\cite{2016_DIAL} suggest a similar network architecture named DIAL. It considers the real world communication limitation, proposing an additional discretize/regularize unit~(DRU). During training, DRU allows messages to be continuous for gradient back propagation, while it discretizes the message in execution. However, this inconsistency between the training and execution phase can lead to unstable behavior.

A novel architecture named BiCNet, also based on the idea of differentiable communication, was introduced in~\cite{2017_bicnet}.
BiCNet applies the bidirectional RNN as not only a local observation memory saver, but also a communication channel. However, it requires a global state as input, limiting its potential application.  
Authors in~\cite{2018_ATOC} take advantage of the attention mechanism and propose an attention-based model called ATOC as an extension of BiCNet. In this model, the attention module acts as an indicator that tells whether communication is required, significantly reducing unnecessary information. Similarly, SchedNet~\cite{2019_schedulecomm} gives each message a weight and only top $k$ messages are broadcasted, thus solving the redundant information issue.

In~\cite{2019_commminimum} emphasis is placed on message representation, and it proposes a network based on communication minimization. This network is designed to maximize the mutual information between messages and actions and to minimize the entropy of messages. Mutual information loss describes how well the messages represent observations, and message entropy indicates how valuable the message is.

In our work also, we will model and train a communication channel for agents in a queueing network. We additionally assume that some dispatchers receive noisy observations about the queue states. The observation encoder is instantiated by variational auto-encoder and the communication channel is modelled as product-of-experts. Similar to SchedNet, we utilize the attention mechanism to cope with redundant information issue.

\section{SYSTEM MODEL}\label{SYSTEM MODEL}
\begin{figure}
    \centering
    \includegraphics{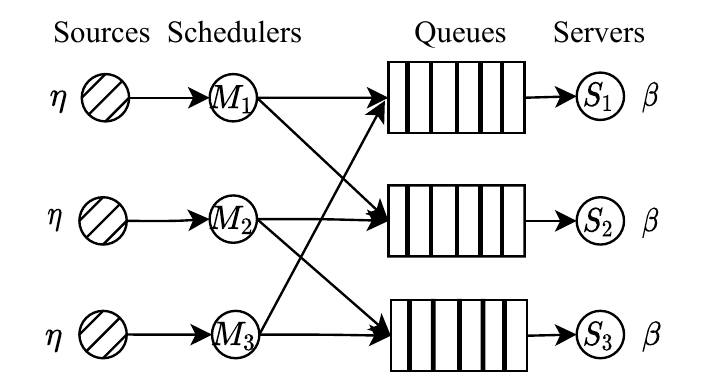}
    \caption{Our system model consisting of schedulers and parallel queues. Jobs arrive to the schedulers, with rate $\eta$, and are then allocated to the accessible queues by each scheduler. Servers serve queues in a FIFO manner, with rate $\beta$.}
    \label{fig:queueing_network}
\end{figure}

We consider a queueing network environment which is composed of $M$ schedulers and $S$ homogenous servers. Each server has its own finite first-in-first-out~(FIFO) queue with a maximum buffer capacity $B$. The service rate of all the servers follows an exponential distribution with rate $\beta$.
Inspired from the scalable \textit{power-of-d} models \cite{mitzenmacher2001power}, each scheduler has access to $d$ out of $S$ queues, where $d \le S$. This means that the schedulers can allocate their arrivals only to these $d$ queues and only observe their states, making the system decentralized.
Each scheduler has its own independent Poisson arrival stream, with rate $\eta$, and to ensure that the system traffic load is not too high or too low we assume $M \eta \approx N \beta$.
After every $\Delta T$ the schedulers send all the jobs that they have received so far to their $d$ accessible queues.
This $\Delta T$ is the synchronization time it takes for the schedulers to get information of their $d$ accessible queues, and is also
the decision epoch for our discrete model.
Please refer to Figure \ref{fig:queueing_network} for an illustrative representation of our system model for~$M=3$,~$S=3$~and~$d=2$.

Since the queues have finite capacity, packets will be dropped if sent to an already full queue and will result in a penalty. 
Due to limited communication capacity and propagation delays, schedulers cannot always observe the true state of the servers they have access to, which means that some observations are delayed (or noisy). This adds partial observability to our multi-agent system. For example, when a $M_1$ scheduler samples the queue states of servers $S_1$ and $S_2$, it may get a delayed observation of $S_1$ and a true observation of $S_2$. However, $M_1$ has no way of knowing which one is delayed. It needs to infer the true queue length and train a policy, based on the observations and the inferred states, that minimizes the total packet drop rate. 

This problem can be modelled as a decentralized partially observable Markov decision process (Dec-POMDP),~\cite{2016_dec-pomdp} defined by the tuple $\left<\mathbb{I}, \mathbb{S}, \{\mathbb{O}\}, O, \mathbb{A},  P, R, b(s) \right>$, where $\mathbb{I}=\left\{1,2,\ldots, M\right\}$ is the set of schedulers, $\mathbb{S}$ is the set of state of all queues, $\{\mathbb{O}\}$ is the set of joint observations of all agents, $O$ is the observation probability function, $\mathbb{A}$ is the set of joint actions, $P$ is the transition probability function, $R$ is the immediate reward function and $b(s)$ is the initial state distribution of all environment.
Since, the schedulers are fully cooperative, they share a global reward:
\begin{align}
    R = \sum_{j=1}^S D_j
\end{align}
where $D_j$ is the number of jobs dropped in each queue. The goal is to minimize the total jobs dropped by all the agents.

In the next section, we describe the methodology we have used to solve the above described Dec-POMDP.

\section{METHODOLOGY}
\begin{figure}
    \centering
    \includegraphics[width=\linewidth]{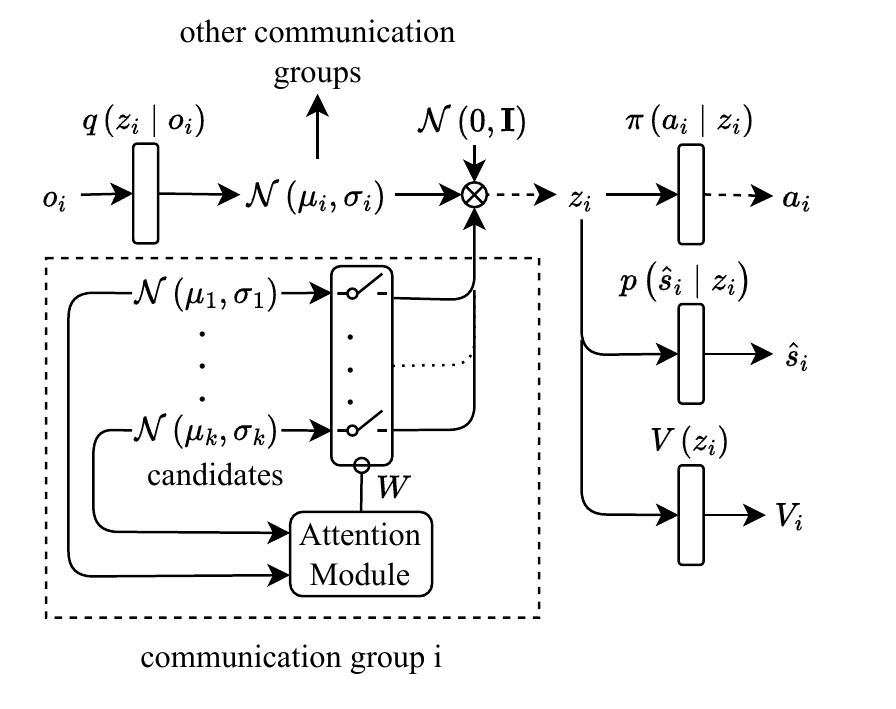}
    \caption{The structure of ATVC model consisting mainly of an encoder and the attention module. The observation received by the scheduler is processed to get an action $a_i$, environment state prediction $\hat{s_i}$ and value $V_i$.}
    \label{fig:model_overview}
\end{figure}

Our proposed attention-based variational communication model, ATVC, applies a multimodal variational autoencoder and an attention mechanism to extract the underlying features and select valuable potential messages. It is realized using the policy gradient method PPO~\cite{2017_PPO}. The overall structure of our model is illustrated in Figure~\ref{fig:model_overview}.

For a scheduler $i$, the encoder $q\left(z_{i} \mid o_{i}\right)$ takes the noisy observations, $o_{i}$, as input and infers a multivariate Gaussian distribution that encodes $o_{i}$ as $\mathcal{N}\left(\mu_i, \sigma_i\right)$, where $\mu$ is the mean and $\sigma$ is the covariance. The $k$ schedulers who have common access of queues with scheduler $i$, called candidates, form a communication group, where the attention module behaves as a selector that determines with which schedulers to communicate. Then the Gaussian distributions of these schedulers and a prior distribution are integrated through product of experts~(PoE)~\cite{2002_PoE}, which helps the individual scheduler $i$ in inferring the true states of the queues. A latent vector $z_i$ is sampled from the integrated distribution and fed into the three heads to generate action $a_i$, environment state prediction $\hat{s_i}$ and value $V_i$.
We will now explain the main components of our ATVC model in detail.

\subsection{Multimodal Variational Autoencoder}
Any natural phenomena generally occurs with several modalities, like a flower with color and odor. Therefore, a model that learns representative features from multiple modalities may have a promising future. In order to approximate the true posterior $p\left(z \mid x_ {1}, \ldots, x_{N}\right)$ conditioning on multiple modalities, \cite{2018_MVAE} proposes the following:

\begin{equation}
    p\left(z \mid x_{1}, \ldots, x_{N}\right) \propto p(z) \prod_{i=1}^{N} q\left(z \mid x_{i}\right),
\end{equation}

where $p(z)$ is the prior belief, typically expressed by a normal distribution, and $q\left(z \mid x_{i}\right)$ is the posterior approximation function, which can be instantiated by multi-layer perceptron neural network~(MLP)~\cite{1974_MLP}, or recurrent neural network~(RNN)~\cite{1997_RNN}. MVAE is a multimodal extension of variational autoencoder, which combines the inferences of different modalities through PoE. Using the PoE, MVAE learns the underlying features much more comprehensively, as compared to the standard variational autoencoder~(VAE)~\cite{2013_VAE}. 

Inspired by the MVAE framework, we represent the partial observations of individual schedulers as modalities. For the same queue, some schedulers get true observations, while some get delayed observations. This could be due to propagation delays or other faults in the channel between the scheduler and queue. These observations, as modalities, are encoded by the encoder $q\left(z_i\mid o_i\right)$ as Gaussian distributions in terms of $\mu_i$ and $\sigma_i$ and then using PoE are aggregated, with the received Gaussian parameters from other candidates. The aggregated Gaussian is the belief over the true state of the queue. After sampling the latent vector $z_i$, the decoder $p\left(s_i\mid z_i\right)$ takes it as input and reconstructs the true state $\hat{s_i}$ of queue $i$.

\begin{figure}
    \centering
    \includegraphics[width=0.45\textwidth]{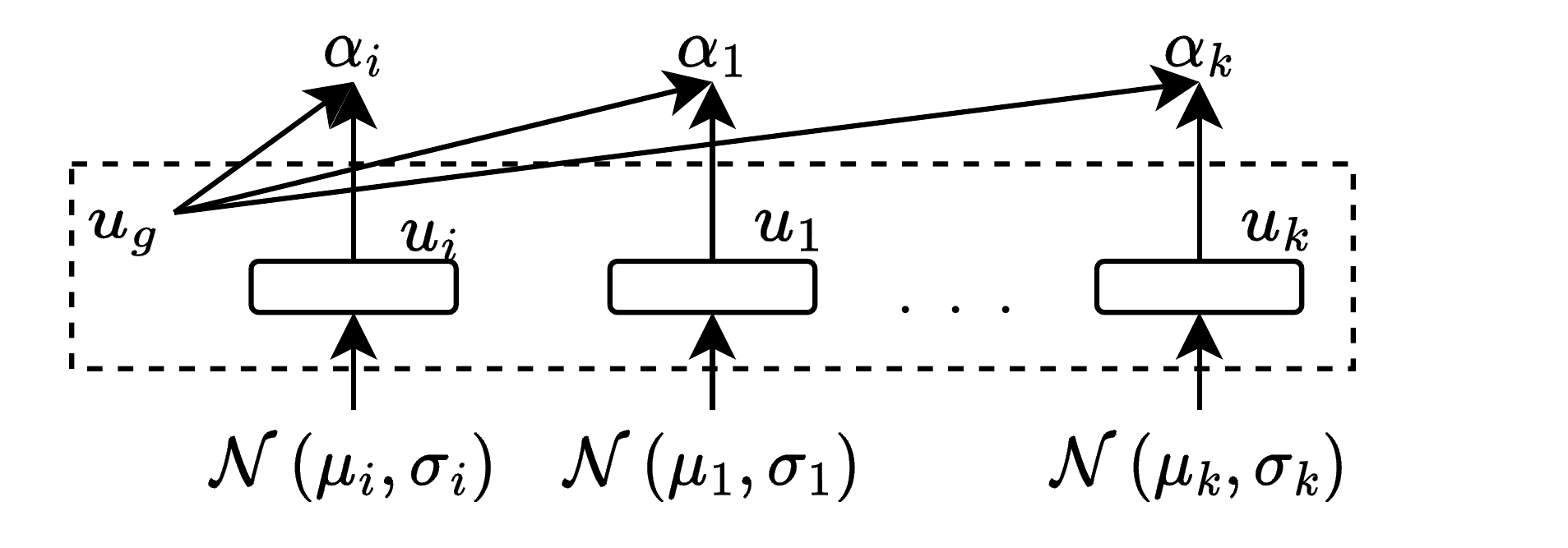}
    \caption{The structure of attention module.}
    \label{fig:attention_unit}
\end{figure}

\subsection{Attention Communication}

When a human wants to understand an article rapidly, it should extract the keywords and deduce the content of the article from those keywords. To imitate this behavior, we introduce the attention mechanism with the purpose of efficient and effective communication. For example, a scheduler $i$ can form a communication group in which all the schedulers have access to the same queue. If we directly aggregate all the inferred Gaussian, the noise in observation may mislead the agents. Therefore, we need an attention module to choose the reliable Gaussian distributions.

The attention architecture we apply is the dot product attention~\cite{2015_dot_attention}, as Figure~\ref{fig:attention_unit} showcases. Here, $u_g$ is a context vector which can be described as a high level embedding of "which distribution is reliable?" as used in memory network~\cite{2015_memory_network}. Notably, $u_g$ is a model parameter that should be jointly trained with other model components. The computation of attention weight is specified as:

\begin{align}
u_i &=\tanh\left(\omega\cdot\left[\mu_i,\sigma_i\right]+b\right) \\
\alpha_{i} &=\frac{\exp \left(u_{i}^{\top} u_{g}\right)}{\sum_{i} \exp \left(u_{i}^{\top} u_{g}\right)} 
\end{align}

That is, we first feed the concatenation of $\mu$ and $\sigma$ into one-layer MLP which has parameters $w$ and $b$, to obtain $u_i$ and then normalize the similarity between $u_i$ and context vector $u_g$ to get the weight $\alpha_i$. 
Then the attention module selects the messages, to integrate, based on the corresponding attention weights.
Messages with a weight greater than a pre-chosen threshold value $\gamma \le 1$ are considered informative, otherwise they are ignored.

\subsection{Training and Execution}
We train our ATVC model under the fashion of centralized learning and decentralized execution. In order to accelerate the convergence process, we are able to use a stricter centralization method, parameter sharing, since we have homogeneous agents in the environment. ATVC consists of $5$ components: encoder $q\left(z_i\mid o_i\right)$, action head $\pi\left(a_i\mid z_i\right)$, value head $V\left(z_i\right)$, decoder $p\left(s_i\mid z_i\right)$ and attention module $\operatorname{att}\left(\mathcal{N}\left(\mu_j,\sigma_j\right),\dots,\mathcal{N}\left(\mu_k,\sigma_k\right)\right)$.  And, we need to learn the model parameters/weights $\omega_q$, $\omega_\pi$, $\omega_V$, $\omega_p$ and $\omega_{\operatorname{att}}$ of each component during training.

We can update the value head using the loss function: ~$\mathcal{L}^{\text{VF}}=\left(V\left(z_{i}\right)-\text{target}\right)^2$. While, the clip surrogate loss~\cite{2017_PPO} for action head can be formulated as:

\begin{align}
    \mathcal{L}^{\pi}_1 &= \frac{\pi\left(a_{i} \mid z_{i}\right)}{\pi_{\text {old }}\left(a_{i} \mid z_{i}\right)} \hat{A}_{i} \nonumber\\
    \mathcal{L}^{\pi}_2 &= \operatorname{clip}\left(\frac{\pi\left(a_{i} \mid z_{i}\right)}{\pi_{\text{old}}\left(a_{i} \mid z_{i}\right)}, 1-\epsilon, 1+\epsilon\right) \hat{A}_{i}\nonumber \\
    \mathcal{L}^{\pi} &=\mathbb{E}\left[\min \left(\mathcal{L}^{\pi}_1, \mathcal{L}^{\pi}_2\right)\right]
\end{align}

where $\hat{A}_{i}$ is the generalized advantage estimation~\cite{2015_GAE} and $\epsilon$ is policy gradient clip factor. We then use the following VAE loss to update the decoder:

\begin{equation}
    \mathcal{L}^{\text{VAE}} =\mathbb{E}_{q\left(z_i \mid \mathcal{O}\right)}\left[\log p\left(\hat{s}_i \mid z_i\right)\right]-\gamma\operatorname{KL}\left[q\left(z_i \mid\mathcal{O}\right), p\left(z_i\right)\right].
\end{equation}

The first term is reconstruct loss, which in our work is a cross entropy loss. The second term is the KL divergence between joint posterior and prior.
Since, ATVC selects messages based on attention weights, $\alpha$, the gradients cannot flow through the attention module. To cope with this issue, we apply a weighted PoE with attention weight during training. The weighted PoE~\cite{2014_gPoE} can be solved in a closed form with mean $\mu =\left(\sum_{i}\mu_{i}\alpha_i\mathrm{T}_{i}\right)\left(\sum_{i} \alpha_i\mathrm{~T}_{i}\right)^{-1}$ and covariance $\sigma =\left(\sum_{i} \alpha_i\mathrm{~T}_{i}\right)^{-1}$, where $T_i = \sigma_i^{-1}$.

The parameters $\omega_q$ and $\omega_{\operatorname{att}}$ for the encoder and the attention module are learned jointly using the PPO loss and VAE loss. Therefore, ATVC is an end-to-end trainable model with a total loss function:

\begin{equation}
    \mathcal{L}^{ATVC} = \mathcal{L}^{PPO} + \kappa\mathcal{L}^{VAE},
\end{equation}
where $\kappa$ is a hyperparameter that scales VAE loss to make training stable.

During execution, the value head and the decoder are discarded. Each scheduler $i$ should infer a Gaussian distribution $\mathcal{N}(\mu_i,\sigma_i)$ based on its partial observation and send the Gaussian distribution as a message candidate to the communication groups of other agents. Meanwhile, the attention module generates attention weights for candidates in the communication group of scheduler $i$. The candidates with a weight greater than $\gamma$ are chosen for PoE. Finally, each scheduler $i$ samples a latent variable from the joint Gaussian and select actions based on it, using a trained PPO policy model.

\section{EXPERIMENTS}
We consider a queueing network environment as mentioned in section~\ref{SYSTEM MODEL}, consisting of $3$ schedulers and $3$ servers. The rest of the environment configuration are given in Table~\ref{tab:env_cfg} and the PPO hyperparameters used are given in Table \ref{tab:ppo_cfg}.

For comparison, we implemented the full communication models like CommNet~\cite{2016_CommNet} and BicNet~\cite{2017_bicnet} and the no communication models, JSQ~(join the shortest queue)~\cite{2015_JSQ} and random policy. JSQ and Random policies always have access to the true queue states. As the simplest model, JSQ has a time complexity of $O(1)$, since schedulers only need to choose the shorter one out of 2 queues. CommNet has $O(MD)$ complexity since they consist of fully connected networks. ATVC and BicNet have the highest time complexity, $O(M^2D)$, due to RNN or attention architecture, where $D$ is the number of neurons.

\begin{table}[h]
    \centering
    \caption{The configuration of the training environment}
    {
    \begin{tabular}{ccc}
        \toprule
        Hyperparameters & Descriptions & Value \\
        \midrule
        $M$ & Number of the schedulers & $3$\\
        $S$ & Number of the servers & $3$\\
        $d$ & Number of accessible queues & $2$\\
        $\eta$ & Packet arrival rate & $0.9$\\
        $\beta$ & Server service rate & $1.0$\\
        $B$ & Maximum queue capacity & $5$\\
        $R$ & Reward for every packet drop& $-1$\\
        $\gamma$ & Attention module weight threshold & $0.3$\\
        \bottomrule
    \end{tabular}}
    \label{tab:env_cfg}
\end{table}

\begin{table}[h]
    \centering
    \caption{The configuration of PPO}
    \begin{tabular}{ccc}
        \toprule
        Hyperparameters & Descriptions & Value \\
        \midrule
        $\lambda$ & Advantage discount factor & $1$\\
        $\nu$ & KL initial coefficient & $0.2$\\
        $B_s$ & Train batch size & $4000$\\
        $B_m$ & Minibatch size & $128$\\
        $l_r$ & Learning rate & $5e^{-5}$\\
        $\epsilon$ & Policy gradient clip factor & $0.3$\\
        $\delta$ & Value clip factor & $10$\\
        \bottomrule
    \end{tabular}
    \label{tab:ppo_cfg}
\end{table}

\begin{figure}
    \centering
    \includegraphics[width=0.95\linewidth]{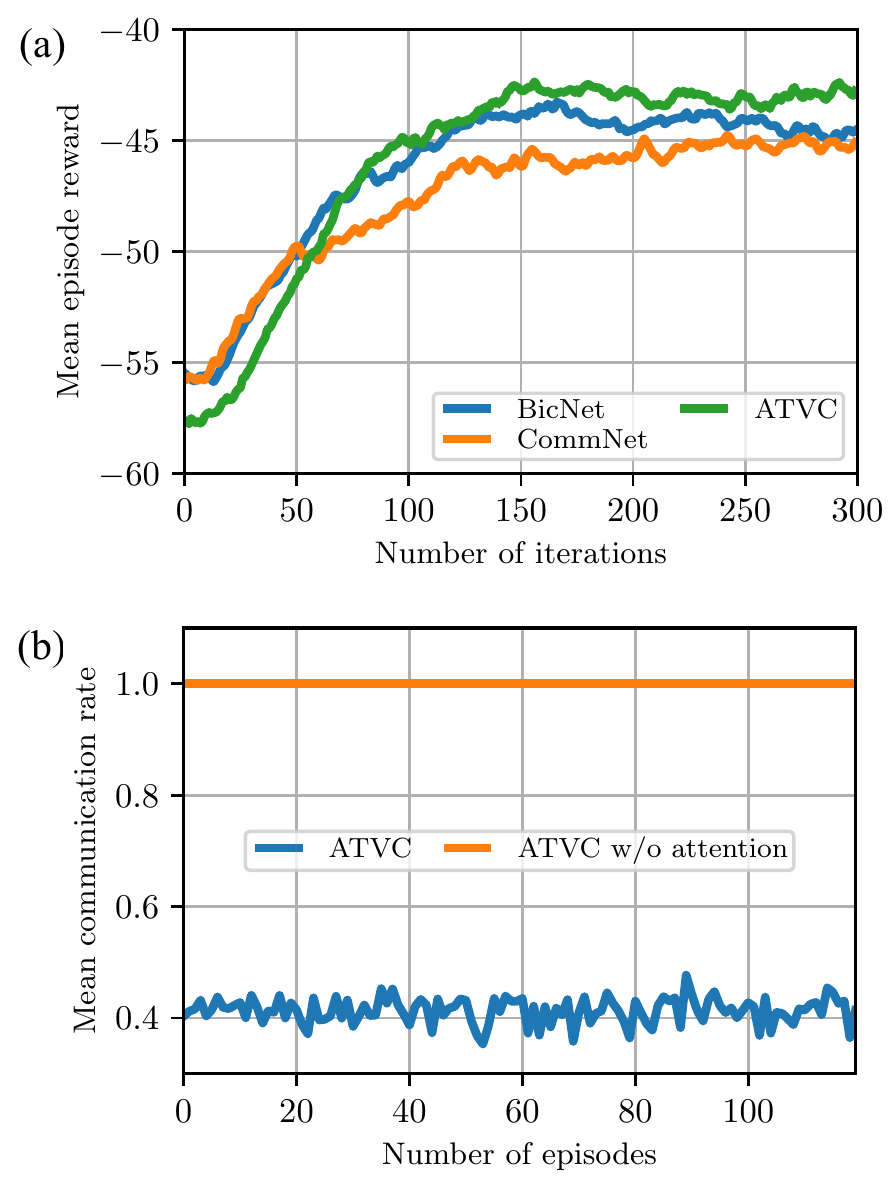}
    \caption{(a) Reward of ATVC against baselines during training and (b) the ratio of schedulers that are selected for the purpose of communication by the attention module of ATVC as compared to the full communication models, where you are always communicating with everyone else. }
    \label{fig:prob_compare}
\end{figure}

\begin{figure}
    \centering
    \includegraphics[width=0.95\linewidth]{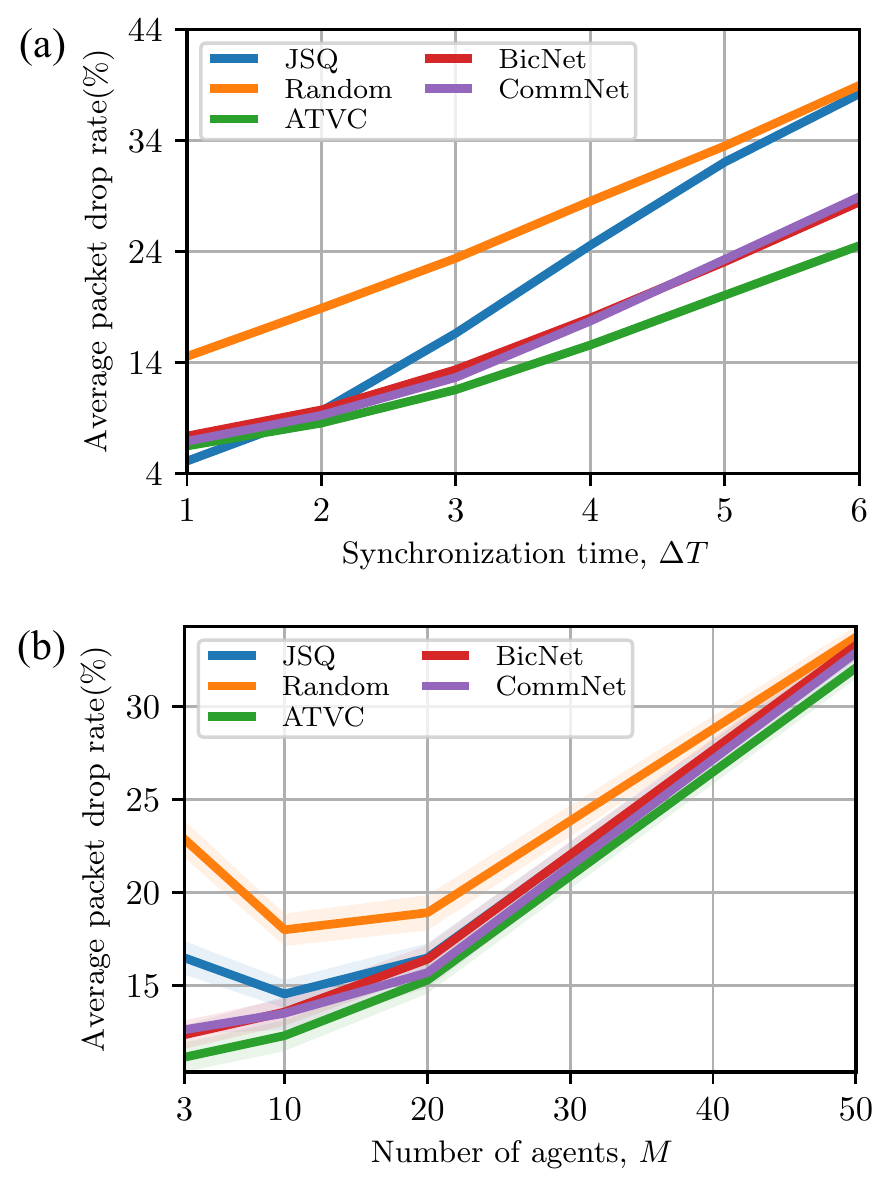}
    \caption{Packet drop rate against (a) the synchronization delay, $\Delta T$ and (b) the number of agents (schedulers), $M$.}
    \label{fig:deltaT_compare}
\end{figure}

Figure~\ref{fig:prob_compare}a showcases the learning curves for $300$ iterations, where $1$ iteration includes $50$ episodes. It can be seen that ATVC converges to only slightly higher reward than CommNet and BicNet. However, the motivation to apply the attention mechanism is to minimize unnecessary messages from candidates. Figure~\ref{fig:prob_compare}b shows how ATVC actually makes communication effective and efficient by greatly reducing it to only $40\%$ as compared to the full communication models, CommNet and BicNet.

Our queueing network is time-discrete and the schedulers route the packets every $\Delta T$s. This $\Delta T$ can be thought of as the delay after which the agents get information of the queue states and get synchronized. Once information on queue states is received, it is maintained by the agents until the decision epoch. To evaluate the robustness of ATVC and other models for increasing $\Delta T$, we recorded the average packet drop rate over $1000$ episodes. Figure \ref{fig:deltaT_compare}a illustrates how the packet drop rate increases with the growth of $\Delta T$, and for JSQ it increases particularly sharply. 
This is because with the increase of $\Delta T$, more packets are accumulated to be sent to the queues at the same time and JSQ, being discrete, is prone to send more than $B$ packets to a single queue, which is beyond the maximum buffer capacity of the queue, resulting in packets being dropped. In comparison with JSQ, other continuous action models are more robust against increasing $\Delta T$, since they can distribute their packets over $d$ queues. Notably, as $\Delta T$ increases, ATVC performs better than other baselines while at $\Delta T=1$ its performance is close to JSQ (which has true state of queues at all times).

In order to investigate the scalability of ATVC and the baselines, we tested them on the different sizes of queueing networks. We only train them once in the queueing network with $3$ schedulers and use the trained model for further evaluations. When switching between different queueing networks, we adjust the arrival rate per scheduler to maintain the constant ratio between the total arrival rate and total service rate, in order to keep the average system utilization always at $90\%$. Figure~\ref{fig:deltaT_compare}b shows that our model is scalable and works well even \textit{with partial information and limited communication}. 
Policies JSQ and Random always have true information of the system, which is not realistic. While, CommNet and BicNet need to communicate all the time with all the agents, resulting in a huge communication overhead.
Receiving all messages from candidates actually deteriorates the load-balancing decisions, because some unnecessary messages are also sent and may mislead the decision-making process.

\begin{figure}
    \centering
        \centering
        \includegraphics[width=0.75\linewidth]{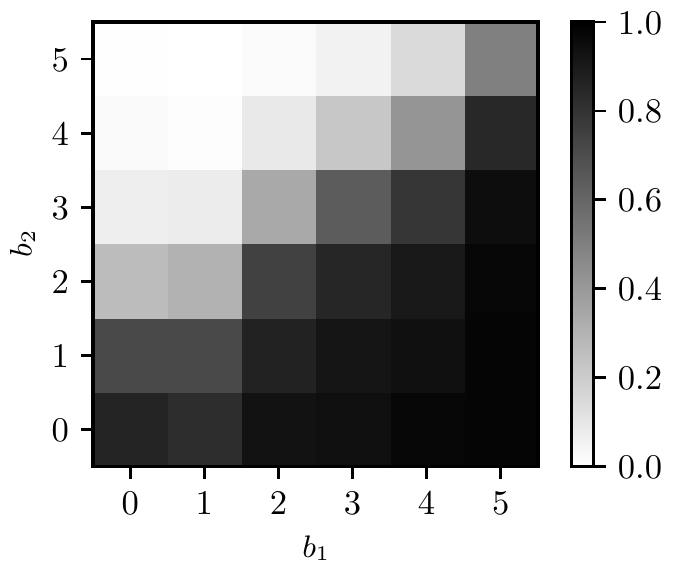}
    \caption{Packet allocation decision of ATVC when a scheduler has access to $2$ queues having states $b_1$ and $b_2$. Heatmap represents the probability with which ATVC dispatches packets to each queue depending on their states. 
    The darker the color, the higher the probability is to send to queue $2$ with buffer capacity $b_2$.
    }
    \label{fig:action_intention}
\end{figure}

We also visualized the packet allocation policy of ATVC, which is the probability of choosing each queue based on the agents' own observation. 
Figure \ref{fig:action_intention} shows that ATVC is able to allocate correctly, i.e., sending more to the shorter queue. Thus indicating that it was able to infer the true state of the queues from received the partial observations.

\section{CONCLUSION}
In this work, we presented and evaluated an attention-based variational communication model, ATVC, in a partially observable queueing network. The framework of ATVC can be divided into $2$ stages. Firstly, it uses the attention module to dynamically select informative messages from candidates with the purpose of queue state inference. Then, it uses MVAE to take the product of these messages for inference. Based on the results of our experiments, ATVC performs better, in terms of average reward, than other communication and non-communication based models, while having partial state information and greatly reduced but efficient communication. We also showed the scalability of ATVC by training it only on a small setup and using it for evaluation of larger setups, in terms of the number of queues and schedulers. As a future work, we would also like to consider heterogeneous servers and other load-balancing policies to compare with.

\bibliographystyle{IEEEtran}
\bibliography{IEEEabrv,refs}

\end{document}